# How does the AI understand what's going on

Dimiter Dobrev
Institute of Mathematics and Informatics
Bulgarian Academy of Sciences
email: d@dobrev.com

Most researchers regard AI as a static function without memory. This is one of the few articles where AI is seen as a device with memory. When we have memory, we can ask ourselves: "Where am I?", and "What is going on?" When we have no memory, we have to assume that we are always in the same place and that the world is always in the same state.

## Introduction

The standard approach in AI is to take a set of positive examples and a set of negative examples. We seek for a function that says "YES" for the positive examples given, and "NO" for the negative examples given. Using the function found, we begin to predict the right answer for examples which we do not know whether are positive or negative.

In essence, the standard approach in AI represents an approximation. What is sought for is an approximation function. It is usually sought for in a given set of functions. For example, in neural networks, the function is sought for in the set of neural networks.

As a typical example of the standard approach to AI we can indicate [1] where a static function covering certain positive and negative examples is sought. In articles like [2], things are a bit different, because there are new examples at every step and at each step the approximation function changes. However, in [2] we are again seeking for a static function without memory, although at each step this function is different. In [2] the idea is that we constantly improve the approximation function, but at each specific moment we have one specific static function.

The only research in which AI is seen as a memory device is the research related to Reinforcement Learning. Even in this area there are two main branches called "Partial Observability" and "Full Observability". Do we need memory when we see everything? Of course not! If we see the current state of the world, we do not need to remember anything from the past because everything we need to know is encoded in what we see. That is, the only branch of AI where AI is considered as a memory device is the Reinforcement Learning with Partial Observability. This article refers exactly to this branch of Computer Science.

Why are we saying that such articles are few? There are many articles in the field of Reinforcement Learning, but they are largely devoted to the Full Observability case. For example [3, 4]. When referring to Partial Observability, it is usually said that there is such an option, but it is not said what we are doing in this case.

In this article we ask the question "What is going on?" If you translate this question in terms of Reinforcement Learning, it will sound like this: "What is the state of the world, we are in at the moment?" If you look at the states of the world as a graph, the question "What is going on?" can be translated as "Where are we? In which vertex of the graph are we in?" If we think of the real world, "What is going on?" means "What is our physical location at the moment and what is the state of the world at this moment?"

To say in what state of the world we are, we must first describe the states of the world. It will never be easy, because we do not see these states completely, but only partially (Partial Observability). For this reason, the description of the states of the world is related to imagining something invisible.

For this purpose we will introduce the concept of "property of the world", representing a set of states of the world. With this concept we will describe the current state of the world. This will happen by the properties of the world that we know are valid or not. In this way, we will determine that the current state is an element of the intersection of a group of sets, i.e. an intersection between properties that are valid at this moment and the complements of the properties that are not valid. If these properties are enough, then the intersection of all these sets will be small enough and the description of the current state of the world will be precise enough. (That is, we do not define the current states with precision to one single state but with precision to a set of states.)

To describe the properties of the world, we will first see what properties we have. All we have are the experimental properties. We will want to describe particular property through what we have. This will happen by stating that from a given experimental property it follows something for the property we are describing. For example, that it follows that the property is true with some probability. The good cases are when it follows that it is true with a 100% probability, or vice versa – it is false to a 100%.

How will we determine the properties? To form a drop of rain we need a speck of dust that serve as a condensation nucleus. Similarly, in order to create a property that is defined for all the states of the world, we need to have the property defined for a relatively small subset. Based on this subset we will make an extrapolation and assume that the probability for the whole set is the same as for this subset.

Who are these condensation nuclei by which we will obtain the required properties of the world? These will be the tests defined in [8] and the corresponding function of the test.

This article is, in its essence, a continuation of [8], but readers shouldn't have necessarily read [8] to grasp the idea of the present article. (Short versions of [8] have been published in [9] and [10]). In [8] the term "Testable state" describes five different things: tests, function of the test, prediction of the function, test property, and test state. Those five different things denoted by a single term can be considered a mistake, though between the first and second there is a bijection. The prediction is also determined completely provided that we've specified the moment it is for (i.e. after how many steps). As to the test property, it is a continuation of the function of the test. This function can be continued in many ways, but we are looking for a continuation that is as natural as possible. The test state depends on the splitting into groups of relative stability. This splitting can be done in many ways, but the meaningful splittings are few.

This article begins with the introduction of three new definitions of Reinforcement Learning and proof of the equivalence of these new definitions with the standard definition. The aim of the new definitions is to help us introduce the concepts of chance, test state and noise. We will give a specific example that justifies the introduction of the new definitions. We will introduce the concepts of event (experiment), we will say what a property of the world is and we will consider the properties defined by experiments. We will introduce the term of test. From it we get the test property. When we have not one, but many groups of relative stability, then instead of a test property we will talk about test state. That means that we will talk about a test that defines not one, but several properties of the world. Finally, we say how to define theories that describe the properties of the world.

Thus we will answer the question "What is going on at the moment?" The answer is a set of properties so that we know whether they are valid at the moment. The next action of our device will not depend solely on what it sees at the moment. It will also depend on what properties of the world are currently valid (that is, on what is going on at the moment). This means that our device will be a device with memory.

## Formulation of the problem

We have a sequence: *action, observation, action, observation* ..., and we want to understand this sequence. Our goal is to predict how this sequence will continue and to choose such actions to achieve the best possible outcome. We will assume that this sequence is not accidental, but that it is determined by the rules of a world in which we are.

The representation of the world is essential because we will try to understand the world, that is, we will try to build a model, and we will seek for this model in a form close to the form we've chosen.

We will look at four possible definitions of the type of world. We will prove that these four definitions are equivalent. The benefit of the three new definitions we will offer is that they will help us in building a model of the world. The second definition will help us add the notion of chance to our model, the third definition will naturally lead us to the notion of Testable Property, and the fourth definition will give us the possibility to add the notion of noise.

**Definition 1.** This is the usual definition of the world in Reinforcement Learning. It is the following: We have a set *S* of the internal states of the world and one of them, $s_0$, is the initial one. How the internal state of the world changes is determined by the function *World*, and what we see at each step is determined by the function *View*. The following applies:

$$s_{i+1} = World(s_i, a_{i+1})$$
$$v_i = View(s_i)$$

Here, actions and observations ($a_i$ and $v_i$) are vectors of scalars with dimensions *n* and *m*, respectively. Each of these scalars will be a finite function with *k* possible values, where *k* is different for the different coordinates of *a* and *v*.

Let's ask whether the function *World* is single-valued or multivalued. By this definition this function is single-valued, that is, the world is determined. With the new definitions this will change.

The next question is whether the function *World* is total or partial. In [8] we argued that we should assume that *World* is a partial function and that cases in which it is not defined will be assumed to be incorrect moves. Again in [8] we have accepted that at each step we will be able to check for each move whether it is correct or incorrect. That is, at each step we will see two things. First, we'll see what the function *View* returns, and second, we'll see which actions are correct and which are incorrect at this moment.

As we have said, this is the usual definition used by most authors (for example [3, 4]), except that other authors usually assume that the function *World* is total and that all moves are correct. We will not prove that the definitions in the case of total and partial function *World* are equivalent because they are not. In the case in which we allow incorrect moves the device gets more information. This can be partially emulated in the case in which there are no incorrect moves, but this emulation will be only partial.

**Definition 2.** With the single-valued function *World* we lack the concept of chance. Let's see how we can add it. If we allow *World* to be multivalued, then the next state of the world will not be definite but will be one of several possible. Okay, but we'd like to say something about the probabilities of these different options. Let's have *k* different options. Let each of them happen with some probability $p_i$. (Now things look like the Markov Decision Process or, in particular, the Partially Observable Markov Decision Process, because here we are dealing with the Partial Observability case.)

So, we can define chance in two possible ways. With the first one the probability is not determined at all, and with the second one it is determined too precisely. Both options are not what we want, so we will choose something in the middle. In the first possible way, the probability is somewhere in the interval *[0, 1]*. In the second way, the probability has a fixed value *p* (i.e. it is in the interval *[p, p]*). In the variant which we will choose the probability will be in an interval *[a, b]*. That is, we will not know the exact probability, but we will know that the probability is at least *a* and less than *b*.

There are two kinds of chance – predictable chance and unpredictable chance. If you roll a dice, the probability of rolling a six is 1/6. This is a predictable chance. When we ask our boss for a salary increase, the answer will be 'YES' or 'NO', but we can not say how likely it will be 'YES'. This is an unpredictable chance. The predictable chance is quite determined because, thanks to the Law of Large Numbers, we know quite well how many the successful attempts will be. The option we chose is a combination of predictable and unpredictable chance. Of course, in order for the definition to be correct, we must ensure that some inequalities are fulfilled. If *World (s,a)* has *k* possible values of probabilities in the intervals *[$a_i$, $b_i$]*, the following inequalities must be fulfilled:

$$a_i \le b_i \qquad \sum_{i=1}^{k} a_i \le 1 \qquad \sum_{i=1}^{k} b_i \ge 1$$

If $Sum = \sum_{i=1}^{k} a_i$ then the following inequalities must also be fulfilled:

$$b_i \leq 1 - Sum + a_i \qquad (1)$$

We will assume that inequality (1) is equality for at least one *i*. We can assume even that it is equality for at least two *i*.

By this, the second definition of the world is complete. We only have to prove that it is equivalent to the first one.

**Note:** We will assume that the function *World* will not be too indeterminate because it would make the world too incomprehensible. For example, supposing that from any state and upon every action, the function *World* with the same probability can move to any other state, this will make the world completely incomprehensible. It would be much more understandable if, in most cases, the function *World* returns a single possibility, and when the possibility is not only one, the possibilities are few and one is much more likely than the others.

**Theorem 1. The** second definition is equivalent to the first.

**Note:** The next proof is technical, so we advise the reader to skip it and take it on trust.

**Proof:** One of the directions is easy, because it is obvious that the first definition is a special case of the second. For the opposite direction, it is necessary for an arbitrary world described in the second definition to build a world equivalent to that described in the first definition.

The idea of the proof is to hide the probability in a natural number. We will have a function *F*, which will calculate the next number from the current random number. Similarly, the pseudo-random numbers in computers are calculated. There, starting from one number (we will start from zero), each subsequent pseudo-random number is obtained by the function *F* from the previous one (i.e. *F (x))*. Function *F* must be sufficiently complex so that we could not guess the next number. Furthermore, *F* must be good, which means that for one particular *Q* all remainders of modulus *Q* are equally probable.

In our case we have two chances and that is why we will add two natural numbers and two functions (*F_good* and *F_bad*). We will like these two functions to be complex enough (noncomputable and not allowing any approximation with a computable function). We will also want the sequence $F^i(0)$ to be without repetition for both functions. Only function *F_good* is to be good for a particular *Q*, but for *F_bad* we will not want anything more. We will use *F_good* to calculate the predictable chance, and *F_bad* to calculate the unpredictable one.

We will define the function *F_good* as follows:

*F_good(0)=0*

*F_good $^{i+1}$(0)= the first unused number of those which give a remainder k on the division by Q, where k ∈[0, Q-1] is selected randomly.*

**Note:** We've defined the function *F_good* through an endless process in which at each step we are making a random choice (for example, pulling balls out of a hat in which there are *Q* balls). In this way we get an noncomputable function. In this article we try to describe a

specific algorithm and everything that is part of this algorithm must be computable. The functions *F_good F_bad* are not part of this algorithm. Their task is only to show the equivalence of two definitions. We only need to show that these two functions exist. We will never build these functions in practice nor calculate them.

**Note:** If for some reason we want to realize the functions *F_good* and *F_bad,* (for example, if we want to build an artificial world in which to test AI), we would use the standard function *Random,* which is embedded in most programming languages. Unlike *F_good, Random* is not perfect, but it's good enough for practical purposes. *Random* does not work with natural numbers but with a 32-bit integer. It is not perfect in the sense that it is complex, but it is not infinitely complex and the next random number can theoretically be guessed (but in practice it cannot). Moreover, the sequence of random numbers is not infinite, but repeats (but after a rather long period). That is, for practical purposes *F_good* can be replaced with *Random.* *F_bad* can also be built using *Random.* We only have to take care of what the function returns. All classes by module *Q* not to be equally probable but to have another probability distribution. From time to time, this probability distribution to change randomly.

After this preparation we are ready to prove the equivalence of the two definitions. Let us have a world according to the second definition, respectively, *S*, $s_0$, and *World.* The new world, which we will build, will have a set of the states $S \times \mathbb{N} \times \mathbb{N}$, initial state *($s_0$, 0, 0),* and function *Big_World,* which is defined as follows:

*Big_World((s, x, y), a) = (s', F_good $^2$(x), F_bad(y) )*

Here *F_good* is square because this function is used twice in the calculation of *s'*.

*s' = World(s, a)*, when *World (s, a)* has one possible value

When *World(s, a)* has *k* possible values, we select one of them as follows:

We divide the interval *[0, 1]* to *k+1* subintervals with lengths from $a_1$ to $a_k$, and the last one with a length of the remainder. We choose a random point in the interval *[0, 1]*. If the point has fallen in some of the first *k* subintervals, *s'* will be equal to the $i^{th}$ of the possible values of *World (s, a)*. Here *i* is the number of the interval in which we have fallen in.

If the point has fallen in the last subinterval, we calculate the coefficients:
$$c_i = \frac{b_i - a_i}{1 - \text{Sum}}$$

We once again take the interval *[0, 1]* and mark the points $c_i$ in it. Thus we divide the interval *[0, 1]* into *k+1* subintervals (some with zero length). We once again select a random point in the interval *[0, 1]* and see in what subinterval it has fallen in. The first subinterval corresponds to *k* possible values of the *World (s, a),* the second corresponds to *k-1* ones, and the last corresponds to 0 possible values. We cannot fall in the last interval because it is with length zero (because we assumed that at least one of the inequalities (1) is equality). We can even assume that the last two intervals are zero. Once we understand how many and what are the possible values of *World (s, a)*, we choose one of them with unpredictable chance. How does this work? If the possible values are *R,* we take the number *(y mod R)+1*. If this number is 1, we take the first of the options, if it is 2, we take the second, and so on.

We didn't say how we choose a random point in the interval *[0, 1]*. For this purpose, we divide the interval *Q* into equal parts. We take the number *(x mod Q)+1* and this will be the number of the interval which we will choose. It does not matter which of the points of this interval we will take because these intervals will be entirely contained in the intervals we look at. For the latter to be true, we will assume the following:

We assume that the numbers *a* and *b* are rational (if irrational, we can make them rational with a small rounding). We will even assume that these numbers are of the type *x/100* (i.e., that they are hundredths). If they are not, we can again make them of the type with a small rounding. (In this case the rounding is acceptable because we are talking about an interval of probability and if the difference is small, it will be felt only after a very large number of steps.) We will further assume that the number *Q*, which we used in the construction of *F_good*, is equal to the least common multiple of the numbers from 1 to 100.

**Note:** At the first selection of a random point we take the number *(x mod Q)+1,* and for the second selection instead of *x* we take *F_good (x),* i.e. we take the next random number.

Thus the equivalence of the first and second definitions is proven.

■

**Definition 3.** The following idea stands behind this definition: What we see can be changed without changing the place where we are. For example, you are in the kitchen and you see that it is painted yellow. The next day you're back in the kitchen, but this time you see it is painted blue.

We will change the function *View* and the states of the world. If the function *View* returns a vector of scalars of dimension *m,* then we will add *m* visible variables to each state. Now the finction *View* will return the values of these *m* variables. Apart from the visible variables, we will also add a *u* number of invisible variables. The idea of invisible variables is that not everything can be seen. For example, if someone is angry with you, you do not see it, but it does matter because it will later affect his actions.

We will introduce the concept of cumulative state and it will consist of the state of the world and the value of all the *|S|.(m+u)* variables. When we want to emphasize that this is not a cumulative state, we will say a standard state.

The function *World* will be defined for the tuple <cumulative state, action> and will return a cumulative state. Again, the function *World* will be multi-valued and will not be total. That is, again, we will allow chance and incorrect moves.

**Note:** A variable can not be bonded with the state to which it is attached. This is especially true for invisible variables. We are asking ourselves whether we should introduce global variables. The answer is that global variables are not needed because every local variable can be considered to be the same for a whole set of states and even for all states.

**Note:** We will assume that the values of the variables will not change too violently because that would make the world too incomprehensible. For example, supposing that the function *World* of each step changes the value of all variables in an absolutely arbitrary way, it would produce a completely incomprehensible world. It would be more comprehensible if most of

the variables are constants and do not change, and when they are no constants – to change relatively rarely and according to clear and simple rules.

**Theorem 2.** The third definition is equivalent to the second one.

**Proof:** To prove the equivalence of the second and third definition is easy. If we assume that all variables are constants (i.e., the function *World* never changes them), we will see that Definition 2 is a partial case of Definition 3, and vice versa, if we look at the cumulative states as standard states, then from the third definition we get the second. The only remark is that variables can be infinitely many (if the states are infinitely many). From there, we get that cumulative states may be too many (continuum), but if we limit ourselves to the achievable states, they will still be finitely or countably many.

∎

**Definition 4.** The next thing we will do is introduce the notion of noise. We will change Definition 3 so that the new function *View* will no longer return the pure value of the visible variables but the value distorted with some noise. To describe the noise, we will need two things – volume and spectrum of the noise. The volume will be a number *Volume* in the range *[0, 1]*. The spectrum of the noise will be a k-tuple $<p_1, ..., p_k>$. For each visible variable we will add more *k+1* invisible variables that will contain the volume and spectrum of the noise of this variable (i.e., we will add more *|S|.m.(k+1)* invisible variables, if *k* is the same for all visible variables). The function *View* will return the value of the visible variable with probability *1- Volume*. With probability *Volume* it will return noise, which will be one of the possible *k* values, each with a probability of the corresponding $p_i$.

Let the cumulative states in Definition 4 be the same as in Definition 3. That is, they will depend on all visible and invisible variables, but will not depend on what the function *View* returns. In other words, here function *View* is not completely determined by the values of the visible variables of the respective state, but also by the values of invisible variables that describe the noise. Also, some predictable chance takes part in the definition of the function.

**Note**: This is how we describe the situation in which at its input the device receives information distorted by some noise. What do we do when the output information is also distorted by noise? We've dealt with the second option in Definition 2 because there we've already introduced the possibility for the same action to have different possible consequences for the world, each one with a different probability.

**Note**: Again we are dealing with a type of Partially observable Markov decision process (POMDP) with the difference that in our case variables have their value, although it is distorted by noise, while with POMDP there is only noise (i.e., for all visible variables the noise is 100% which means Volume = 1). The only thing that distinguished the various states in POMDP is that they have different noise spectra.

**Note:** We will assume that the world is not too noisy, because if everywhere the noise level is maximum (i.e. one) and everywhere the spectrum of the noise is the same, then such a world would be completely incomprehensible. It would be more comprehensible if the noise level is zero or close to zero.

**Theorem 3.** The fourth definition is equivalent to the third.

**Proof:** It is clear that Definition 3 is a special case of Definition 4. To every world of Definition 3 we can add noise whose level is zero everywhere and we will get a world equivalent to it by Definition 4.

Let's do the opposite. Let's take a world by Definition 4. For each of the cumulative states in this world we will see how many possible outputs the function *View* can return (in principle, the possible output is only one, but because of the noise we can have many possible outputs). For each of these possible outputs, we will make a new state in which the visible variables have exactly the value of the possible output.

**Note**: From every cumulative state we make many new standard states. What we will get will be a world by Definition 3 (and even by Definition 2 because the visible variables will be constants). But do we lose any information by this? Do we lose the value of the visible and invisible variables? No, because this information is coded in the new state we create. It reflects the value of all variables of all states (not just of the variables of the current standard state).

How will we define the function *World* of the new world? If between two cumulative states there is a connection when the action *action* takes place and there is a probability that this connection is likely to take place in the interval *[a, b]*, then each of these two cumulative states has been replaced by many standard states and between each state from the left ones and each state from the right ones there will still be a connection when the action *action* takes place. The difference will be only in the probability interval. It won't be *[a, b]* but *[a.p, b.p]*, where *p* is the probability that it is exactly this output obtained on the right. That is, no matter what state you started from on the left, they all behave the same way because they are the same according to the future. They differ only in the present (function *View*) and they are distinct only because of the noise.

■

# Example

We will give a specific example that will show us the benefits of Definitions 2, 3 and 4. The example will be similar to the one we gave in [6, 7]. The difference will be that here as an example we will use the chess game, whereas in [6, 7] we used the Tic-Tac-Toe game. The main differences are two:

**1.** In chess there are 64 squares, while Tic-Tac-Toe has only nine.

**2.** In chess we will give the commands "pick up the piece" and "put down the piece", while in Tic-Tac -Toe instead of these two commands we have only "put a cross."

Let's have a world in which we play chess against an imaginary opponent. We will not see the whole board with all the pieces. Instead, we'll see only one square and the piece on this square. Let us remind that here we are dealing with the case of Partial Observability. If we were able to see the whole board, we would deal with the case of Full Observability.

However, seeing only one square will not be a problem, because we will be able to move our eyes, i.e. change the square we see and thus look around the entire chessboard.

Our action will be 3-tuple consisting of <horizontal, vertical, command> where:

horizontal∈{Left, Right, Nothing}
vertical∈{Up, Down, Nothing}
command ∈{"pick up the piece", "put down the piece", New Game, Nothing}

The function *View* will return 3-tuple <chessman, color, immediate_reward>
chessman∈{Pawn, Knight, Bishop, Rook, Queen, King, Nothing}
color ∈{Black, White, Nothing}
immediate_reward∈{-1, 0, 1, Nothing}

Our action will allow us to move our eyes on the chessboard – horizontally, vertically, and even diagonally (i.e. simultaneously horizontally and vertically). We will need patience, because we will move only one square per step.

In addition to being able to look around the board, we will need to be able to move the pieces. To do so, we have two commands: "pick up the piece", i.e. pick up the piece you are looking at at the moment. The other command is "put down the piece", i.e. drop the piece you picked up and put it on the square you looking at at the moment. Of course, you will not see which piece you picked up, but we hope you remember it.

For our convenience, we will assume that we always play with the white pieces and that when we move the piece (i.e., we put down the picked up figure) the imaginary opponent will immediately (exactly on the same step) make his move, i.e. on the next step one of the black pieces will be in another place. When the game ends, immediate_reward will be 1, -1 or 0, depending on whether we have won or lost, or the game has ended in a draw. In the rest of the cases immediate_ reward will be Nothing. We will assume that when the game ends, we will not start the next one immediately, but we will have time to look at the board and find out why we lost. When we are ready for the next game, we call the command "New Game" and the pieces are arranged for a new game.

Will there be wrong moves? Yes, when we are in the left column, we will not be able to move to the left. When we are looking at a black piece or an empty square we will not have the right to pick up a piece. If we have already picked up a figure, we will not have the right to pick up another one until we have put down the first one.

Let us describe this world in the terms of Definition 1. The set of the internal states will consist of three things (of 3-tuples). The first will be the position of the board (the possible positions are many), the second will be the coordinates of the eye (64 possibilities), the third will be the coordinates of the picked up piece (65 options – one extra for the case when we have not picked up anything). In order to make the things to be determined, we will assume that the imagined opponent is determined, i.e. at one and the same position he will always play the same move. (Most programs that play chess are determined opponent.) In this case, it is clear how to define the functions *World* and *View*.

What are we to do if we want the imagined opponent not to be determined? For example, it may be a person or even a group of people (who are changing and alternating to play against us). The person, and especially the group of people, is an undetermined opponent. It is natural that at certain position some moves are more probable and others less probable, but it is not possible to tell exactly what the probability of choosing a certain move is. In this case, Definition 2 is most natural.

The states of the world will still be the same, but the function *World* will now be multi-valued. If we want to go back to Definition 1 but preserve the indeterminacy of the imaginary opponent, we will have to make a complex construction of the type we did in the proof of Theorem 1. This would increase the number of internal states of the world. They are too many even now, but they are finitely many, while in the example above they will become infinitely many.

How would this example look like with Definition 3? In this case, the states will be only 64 (as many as the squares on the board.) Each state will have three visible variables (chessman, color, immediate _ reward). The third visible variable will be common for all 64 squares. The fact that the third variable is common to all squares is not said in any way. The device that tries to understand the world will have to detect this fact alone. Of course, this fact is not difficult to detect. Much more difficult to detect is the fact that the first two visible variables depend on the state of the world and are different for each square.

In this case, we will not be able to go only with the visible variables. We need somewhere to remember which piece we've picked up. This fact is not visible in any square. So we'll add an invisible variable to each square. When we pick up the piece, it will disappear from the visible variables and will appear in the invisible variable of the square from which we have picked it up.

Thus we've constructed a world with 64 states and four variables to each state (three visible and one invisible). The number of achievable cumulative states in the world under Definition 3 is just as much as the achievable states under Definition 2. However, the world with 64 states seems much simpler and more understandable, justifying the introduction of Definition 3.

Let us now present the world in the terms of Definition 4. There is no noise in this world and there is no point in presenting it by Definition 4. To make such a presentation necessary, let's add some noise.

We will assume that the white is very dark and the black is very bright and that it is likely to mess up the color of the piece. We will present this in the following way. To the visible variable "color" we will add noise with some volume and spectrum: 50% Black, 50% White, Nothing 0%. When the square is empty, we will assume that the noise is with *Volume=0*.

Next we will assume that the pieces Pawn and Bishop are very much alike and we may mess up them. To present this, we will add noise to the visible variable piece, which noise will have a non-zero value when the figure is a Pawn or a Bishop. For the other pieces let the noise be zero. The spectrum will be Pawn 50%, Bishop 50% and 0% for the other cases.

Now let's assume the King is too feminine and sometimes we mess up him with a Queen, but not the other way around, i.e. we never mess up the Queen with a King. We will present this by adding a little noise when the piece is a King with a spectrum: Queen on 100%.

Thus we saw that we can describe a rather complex world in a very comprehensive way. This example justifies the introduction of Definitions 2, 3 and 4.

## Event or experiment

An event would be when something happened, and an experiment when we did something. Of course, for every event we may have tried to cause it or prevent it. Therefore, we believe that with every event we have some involvement. That's why we will not distinguish between an event and an experiment and will accept that these two words are synonymous.

We want to define the concept of experiment (event). To do this first we will say what history is and what local history around the moment $q$ is.

**Definition**: History is the sequence of actions and observations $a_1, v_1, ... , a_{t-1}, v_{t-1}$, where $t$ is the current moment.

**Note:** We will not tell in which world a given history has happened because we will assume that it is the world that we have to understand. It does not matter by what definition this world is defined because all four definitions are equivalent (we will use mostly Definitions 3 and 4 because they are most convenient to work with).

**Note:** What is one step from the history? Should it be <action, observation> or vice versa? We've decided it would be <action, observation> because the moment in which we are thinking is the moment before the action. We are not thinking in the moment after the action. In this moment we just wait for the observation to take place. For this reason, the history begins with $a_1$. Our device will have to make the first action blindly because it would still not have seen anything. So we can choose the first action randomly. We will choose it to be the vector of zeros (we will denote the zero by Nothing – see [6]). We have said that the world starts from $s_0$, but we do not see what happens in $s_0$. Therefore, the world begins actually from $s_1$.

**Definition:** Local history around the moment $q$ is a subsequence obtained from a history, where from the number of each index we've subtracted $q$.

The representation of the local history is: $a_{-k}, v_{-k}, ... , a_0, v_0, ... , a_s, v_s$. We've obtained it from one particular history by taking the step $a_q, v_q$ and adding the last $k$ steps before it and the next $s$ steps after it. Once we subtract $q$ from the number of each index, step $a_q, v_q$ becomes $a_0, v_0$.

We will look at the local history as a sequence of letters (that is, as a word). We will present this word as a concatenation of two words *past.future*. Here *past* ends with $a_0, v_0$ and the *future* is from there onwards. That is, the present is part of the past, because it has already happened.

We want to define the concept of experiment (event) as a Boolean function that is monotonous (that is, if the event has happened in one particular local history and if the local history is continued, it still would have happened). We also want this Boolean function to be computable.

**Definition A:** Experiment is a Boolean function defined on local histories *past.future,* which is defined by two decidable languages $L_1$ and $L_2$ and is true exactly when $\exists u_1, u_2$ such that $u_i \in L_i$ and $u_1$ is the end of *past* and $u_2$ is the beginning of the *future.*

**Note**: We shouldn't necessarily become aware of the event at the moment it happens. We can become aware of it later (when it goes in the news). This is the reason why the event depends not only on past but also on the future. We might not need to be aware of the future and not even the entire past. That is, we can understand that the event will happen before it has happened (for example, a few steps before it happens).

In this definition of an event, we miss some events that we need to count from the day of birth. For example, the event "Today is Monday" is an event of this kind. If you do not want to miss these events, we will have to change our definition A as follows:

**Definition B:** The same as Definition A with the difference that you want local history to start from the beginning (from $a_1, v_1$) and $u_1$ will not only be an end of *past* but will be equal to *past.*

In [8] we discussed dependencies without memory (i.e. events, by Definition A, in which the lengths of $u_1$ and $u_2$ are restricted). In [8] we discussed the dependencies with memory (i.e. events, by Definition A, where $L_1$ and $L_2$ are regular languages.)

**Note**: Regular languages can be described as words starting with something, containing something, ending in something and in which something has happened *m mod n* times. Definition A tells us that *past* must end in something or contain something. Definition B adds the cases when *past* must begin with something and cases when something has happened *m mod n* times. However, we are not interested in cases when *past* begins with something or contains something. We are not interested in these cases because although we are theoretically looking at many possible histories, in fact, the history is just one (our history). This single history has either begun with something or not. In this history, something has either happened or has not happened. It would be interesting, if something happened recently (for example, no more than 3 steps ago). This would create an event that changes its value over time. Events that are constant are not interesting to us.

# Experimental properties

Once we've said what an experiment is, we are ready to define experimental property. Let us first define property and local history around a state.

**Definition:** Property is a set of cumulative states of the world. At one particular moment, a property will be valid if the corresponding cumulative state is an element of the property (i.e., of the set of the cumulative states).

**Note:** In the Definitions 1 and 2 there is no difference between the standard and cumulative state. In this case, the property is simply a set of states.

**Note:** When we discuss property, we will usually mean not the set but its characteristic function. We will talk about partial properties (whose characteristic function is partial) and the continuation of the partial property to total.

**Definition:** Local history around the state *s* is a local history around a particular moment *q*, which is obtained from a history in which the corresponding state $s_q$ is exactly the state *s*.

That is, local history around *s* is a history that tells us how we have gone through *s*.

**Note:** We can have many different histories around the state *s* because the past and the future are not defined unambiguously. The ambiguity of the future comes from the fact that we do not know which of the possible actions we will choose. In Definitions 2, 3 and 4 to this ambiguity we've added the ambiguity of chance. As for the past, it is also ambiguous as there may be many different states that after an action could lead to the state *s*. Of course, we can assume that each successive state is brand new. That is, assume that the states do not repeat themselves. However, we prefer to assume that there are no unnecessary states in our world (i.e., if two states are equivalent according to the future and to the present, we have merged these two states into one). That is, we will assume that the states can be repeated. Even cumulative states can be repeated, and standard ones are often repeated.

Each experiment defines one property in the following way:

**Definition:** Experimental property is the set of cumulative states *s*, such that for *s* there is a local history around *s*, such that in this local history the experiment has happened (i.e. the experiment was carried out in this local history).

Each experiment defines a property, but that property is not recognizable but semi-recognizable. That is, if the experiment is performed, the property is valid (at that moment). We can not say the opposite. If the experiment has not been performed, we can not say that the property is not valid because in another development of the past and the future maybe the experiment would be possible. If we have noise (Definition 4) then even the present may have another development because of the noise.

Each experiment divides the set of cumulative states into two parts (those that the experiment may be performed around and those for which this can not happen). When the experiment has been conducted around a cumulative state, this means that it is one of the states of the property. Not all states of the property are equally probable. Some are more probable, others are almost improbable. We can not say what the probability of the corresponding cumulative state is, but we hope that gathering statistics for a particular experiment will indirectly take account of these probabilities.

Experimental properties are the best we have. If we want to define a property, we will have to describe it through the experimental properties. This description will be made using the statistics we have collected. In [8] we saw how to collect statistics for dependencies without memory. Also in [8] we discussed even events with memory.

# What is a test?

The drawback of experimental properties is that they are semi-recognizable. We want to add properties that are recognizable (they may not be recognizable at any moment, but there should be moments when we can say whether the property is valid or not).

For this purpose we will introduce the term "test".

**Definition:** Test is an experiment with a result. The result is a Boolean function that is defined always when the experiment has been performed and which does not depend on the way the experiment has been performed. We will call to the experiment the condition of the test.

The idea is, each time an experiment is conducted to have a result and this result to be "YES" or "NO". We do not want the result to depend on the past and the future, because there are many possible developments for the past and for the future. So let us assume that the result depends only on the present.

Present is what we see at the moment (the moment of interest, not the current moment because the current moment is one, but we are interested in all moments). That is, the present is $v_q$. We see this, but we see something more. We see which of the actions are correct at this moment and which are not. To the vector $v_q$ we can add Boolean variables, one for each action. These variables will be visible. The value of each of them will tell us whether the action is correct at this moment.

**Note:** When writing a theoretical article we choose a structure that is easy to describe. This article is a practical one and therefore we will choose a structure, which is most suitable for realization. Therefore, instead of describing the possible moves, we will describe the cumulative of moves (see [8]). The idea is not to play with the individual moves, but to operate with whole sets of moves. We will see two Boolean variables for each cumulative move. This variable will tell *all* and *nobody*.

In [8] we've presented some arguments showing that we can assume that the result of the experiment has the form of $x_i = constant$. Here $x_i$ is one of the visible variables and *constant* is one of the possible values of this variable.

**Note:** In [8] we decided that we will not necessarily try all possible moves in order to see which ones are correct. In the example we gave the possible moves are 36. If we try them all at each step, it would be annoying. Therefore, if the visible variable is one of those who we call *all* and *nobody,* then we will assume that the condition of the test guarantees that the necessary moves have been tried. If the value of the variable is true, then all the moves of the group should have been tried. If the value is false, then at least one move from the group must have been tried and it should be such a move that shows that the variable is not true.

# Test functions

Each test determines one function on local histories.

**Definition:** Function of the test is the function defined in the moments in which the condition happened. The value of this function will be equal to the result of the test.

We want to continue the test function so that we have some prediction of the moments when it is not defined.

**Definition:** Theory of the test is a function that for each local history returns two numbers (prediction and confidence). We will assume that when the function of the test is defined, the theory returns the value of the function as a prediction with a confidence of one. In other cases, the confidence will be less than one.

The prediction is usually zero or one, but it may be between these two values. In this case, we have a prediction of the result with some probability.

Each test function defines a partial property. We will call it the smallest property of the test. The set in which this property is defined is the experimental property (here the experiment is the condition of the test). The value of the partial property is the value of the test result.

**Note:** The smallest property of the test is a generalization of the test function, because it is defined for particular moments for which the function is not defined.

Can we continue the smallest property of the test to a total property? Yes, we can, but we can do this in many different ways. The goal is to continue it in a natural way so that the resulting property actually describes the world.

**Definition:** Property of the test is each continuation for the smallest property of the test.

In [8] we discussed the example where the test is "Is the door locked?" At the moments when we've conducted the experiment, we know the answer. That is, we know the test function. We can correctly define what the correct answer would be for the states around which the test can be performed (although we will not know this answer, if we have not conducted the test). Is there a property of the world that describes the outcome of this test? Let us have the property "the door is locked" and let this property is changing by some rules. It seems like we only need to determine that property so we can predict the outcome of this test. In fact, in our world we may have many different doors and then it is not appropriate to describe the state of the world with just one property. Of course, we could say: "The door to which we are located is locked." This is a property that is always defined except when we are not to any door. However, a better description of the world would be if we introduce many properties (one for each door). So we come to the definition of a test state:

**Definition:** Let's assume we've split the set of cumulative states of the world into groups that we call groups of relative stability. The test state will be a Boolean function that is defined for each moment and for each of these groups.

Roughly speaking, the test state tells us which doors are locked and which doors are unlocked at the moment.

**Definition:** Theory of the test state is a function that for each local history and for each group of relative stability returns two numbers (prediction and confidence).

**Note:** We may have a group of relative stability in which the test is impossible. Then what to do? First possibility is to assume that the theory of the test state will give a prediction of this group with a confidence of zero. Second possibility is to make prediction anyway. Look at the open problem below.

How can we obtain the test theory from the theory of the test state? That is, how the fact that we have an idea about which doors are locked and which doors are unlocked will give us an idea of whether the door in front of us is locked? First, we need to answer the question of which door we are at (i.e., in which of the groups of relative stability we are in). Then we will take the prediction for this door that the theory of the test state gives us.

The groups of relative stability will be represented by the states of a finite automaton. If the finite automaton is deterministic, we will know exactly which group we are in. If not, we will know only approximately. If the automaton is deterministic, the test state can be expressed by several tests and the properties of these tests. The conditions of these tests will be the condition of the test plus the fact that we are in the corresponding state of the automaton. That's why we prefer the automaton to be deterministic. Otherwise, the condition that we are in the respective state of the automaton will not be a computable function.

## Test states

We gave a formal definition of what a test state is. Let's say what the relationship between test states and Definition 3 is.

If the condition of the test is the universally valid condition, then the test states are exactly the visible variables. This is not entirely accurate, because the tests are Boolean and the visible variables have *k* possible values. To fix this inaccuracy, we will temporarily assume that the visible variables are also Boolean.

Let's have a test whose condition is the universally valid one and which returns the value of the first visible variable. Then the state of this test will be the first visible variable. Of course, this is not one variable because every standard state has its first visible variable. We are talking about many variables (perhaps even infinitely many). Many of these variables may be equal. They may even all be equal. Then the test state will not be composed of all of these variables. Instead of this we will take one variable from each group of equal variables.

That is, we have seen that if we have a world by Definition 3, its visible variables are test states. Let's do the opposite. Let's take the world by Definition 2 and a test whose result is the first visible variable. Let's build a world by Definition 3, whose first visible variable is exactly the test state of this test.

For this purpose, we will split the states of the world into groups of relative stability. Which splitting are we to choose? Whichever we choose will do the job, but it is good to make the splitting so that it actually corresponds to the world and that the groups are really groups of relative stability.

Let us now assume that each group of relative stability is one standard state. We'll add more invisible variables, if necessary, to store in them all the information we have about the cumulative state. We will define the functions *View* and *World* in a way that the resulting world is equivalent to the one from which we've started.

**Note 1:** We can represent a group of relative stability through several standard conditions. This splitting is useful, though not necessary. With this splitting the world can become much simpler. For example, instead of thinking that all permanently locked doors are in one standard state, it might be simpler if we have different states corresponding to different permanently locked doors.

What would happen if we choose an inappropriate splitting? Let us imagine that we have split the number of doors into metal and wooden ones. We assume that if one of the metal doors is locked, then all the metal doors are locked and likewise with the wooden doors. Then we see a locked metal door and we conclude that all metal doors are locked at the moment. The next moment we see an unlocked metal door. We decide that they are all unlocked now. Is it possible it really be so? It is possible and it cannot be confirmed or rejected experimentally. However, if the splitting is not adequate, the groups of relative stability will be very unstable.

## Examples of test states

Let's take the world we described in the example (the game of chess). Let's suppose that the description of this world is by Definition 2. We will discuss two tests and see how through their test states the world can be split into groups of relative stability and thereby the world to be presented by Definition 3.

The first test will be "I see a white piece". The condition of the test will be the universally valid condition (that is, when do I check whether I see a white piece. I always check.) The result of the test will be color=white.

What are the groups of relative stability? These will be the squares on the chessboard. The test property will be "There is a white piece in the square we are looking at". The test state will be the position of the chessboard. Well, not exactly the position but that on which squares there is a white piece. The test state will not contain information about where the black pieces are or about which exactly is the white piece that is in the square.

We received a world with 64 standard states. The visible variables are clear. We need to add invisible variables in which to store the information about the cumulative state that has not stored in the visible variables. We can make this in the same way as we already did.

The second test will be "If I see a white piece, I can pick it up."

The result of the test will be "Pick up the piece is a correct move." We should note that "pick up the piece" is not a single move, but a whole group of moves (i.e. a cumulative move) because by picking up the piece we can simultaneously move horizontally or vertically. That is, the result of the test is "In the group of moves <*, *, pick up the piece> There is at least one correct move." We cannot have all moves of the group to be correct, because a move may be incorrect because of the movement (for example, if we are in the left column and try to move to the left). This group has a visible variable *nobody,* which has to be *false*. Formally, it will look like this: *nobody(<*, *, "pick up the piece">)=false*. Here is given for which cumulative move *nobody* is, because different cumulative moves have different variables *nobody*.

The condition of the test will be "I see a white piece" (i.e., color=white). We must add to the condition of the test that we have tried the moves from the group. If *nobody* is *true*, we've tried all the moves, if *nobody* is *false*, we've tried at least one move, but such one which is correct.

Here the groups of relative stability are two: states for which the test will always return *true*, and those for which it will always return *false*. We will have *false* when we've picked up another piece and have not put it down yet and when the game is over and we have not played "New Game" yet in order to begin a new one.

How will we present the world by Definition 3? As noted in Note 1, a group of relative stability can be represented by more than one standard state. Here we will present the group in which the test always returns *false* by two standard states (when we picked up a piece and when the game is over). Let's note that there is no way to pick up a piece if the game is over.

We got a world with three standard states. The visible variables are clear. We need to add invisible variables in which to store all the information for the cumulative state. In other words, we need to store the position of the chessboard (for this we will need 64 variables or 2.64 if we are wasteful). We also need to store the coordinates of the square that we are looking at, and if we've picked up a piece – where we've picked it up from, and which this piece it is.

## How do we find the theories?

What is the difference between the theory of the test and the theory of the test state? In the first case we assume that we have only one group of relative stability and that the test determines one property. In the second case, we assume that we have many groups of relative stability and that our test determines many properties of the world (one for each group).

We will describe how we determine the theory of the test. (The theory of the test state is determined in an analogous way.)

We have collected statistics for specific experiments. When both the experiment and the test are performed, we count how many times the test has returned "YES" and how many times it has returned "NO". Let these be the numbers *n* and *m*. Then the prediction given by this experiment is $\frac{n}{n+m}$, and the confidence depends on *n+m*. At some point, many experiments have been conducted, each of which gives us some kind of prediction with some confidence. Here we will not discuss how to calculate the overall prediction and overall confidence.

**Note:** When a test state is distorted with noise (Definition 4), then we can not find rules that give an exact prediction (i.e. the prediction to be an integer). Each prediction, which is with a large enough certainty, will be approximate (i.e. it will be some probability *p* such that *0<p<1)*. Conversely, if the prediction is approximate, there is no way to know if it's because the result of the test is distorted by noise or because the condition of the test is such that it gives a rough prediction. If all predictions are approximate, we may assume that we have some noise or that we have not yet found a rule which will give us an exact prediction.

Besides experiments, we will predict the property of the test based on the assumption that this property is stable. We will assume that we have gathered statistics on how stable this property is. Based on this, we will assume that once we have checked the property (we have done the test), the value of the property is still the same for some time. The confidence of this assumption will decrease over time (i.e., the more steps goes after the test, the less we count on the assumption that the value is the same).

The next level is to make the assumption that we are not alone in the world. That there are other agents in the world and that this agents change the test state at their own discretion (see [8]).

How do we define groups of relative stability to define a test state that is meaningful and adequate. Once again we will use statistics. In fact, this is the task of finding a finite automaton that can meaningfully split the states of the world into groups. This task will also remain outside of the topics discussed in this article.

**Open question:** Sometimes the test property can not be tested. There are two such cases. The first is when a test property does not make sense. The second case is when it makes sense, but we can not test it directly, and therefore we must assess its value indirectly. The question is how to distinguish these two cases? For example, "Our house burnt down" and the case "My hands are tied". In the first case the test property "The door is unlocked" does not make any sense. In the second case the property makes sense, but we can not test it directly because our hands are tied. We can assess the property indirectly by using a rule of the type "Today is Monday, and on Monday the door is always unlocked". That would be a proper reasoning in the second case but not in the first.

## Conclusion

This article began with the claim to be different and offer for AI something more than an approximation. However, here again the approximation method is used. When we have a test function and we want to continue it to a test property, then what we do in practice is an approximation. However, there are differences. Most authors search for an approximating function that should be the solution (i.e. it must be AI). Here we seek not one but many approximating functions (one for each property). These functions are not the AI. Their only task is to assist the device to understand what is going on at the moment.

**Note:** So to speak, if we had Full Observability and if we were able to see everything, these approximations would not have been needed at all. Although the search for test properties is directed to the case of Partial Observability, it would be helpful for us even in the case of Full Observability. The problem with Full Observability is that in this case we have too much information and it is very difficult to distinguish the essential from the inessential. Let's forget what we see and focus only on test properties that we have found. We will choose these test properties that are interesting and this will be the essential information that we will need.

As we have said, the approximation is not the whole solution, but only part of the solution. Before that, we need to collect statistics to have the basis on which to approximate. Here we should mention dependencies with and without memory, which we've discussed in [8]. Then we make an approximation based on the experiments and the statistics we have collected about them. We also use the assumption for the stability of the test states (this can also be regarded as an approximation). The next method we use to determine the value of test states is the assumption that there are other agents in the same world apart from us and that they are changing these test states in order to help us or screw us. This approach can no longer be called approximation, at least because there is no formula to compute it by. That is, the description in this article is more than just an algorithm for approximation.

Even after finding the test states the problem is not yet solved, because what remains is to plan the future actions to obtain maximum rewards.

**Note:** Here we set the task to describe a particular algorithm and it is the algorithm of the thinking machine. We are not asking ourselves what AI is or the question what the definition of AI is. These questions were discussed in [5] as well as in other earlier articles. That's why in [5] we use the classic definition of Reinforcement Learning, while in this article we have offered three equivalent definitions. When we want to give a theoretical definition of AI, then the classic definition of Reinforcement Learning us enough, but when we want to describe AI in sufficient detail to bring it to a realization, then we need to change the definition so that it is easier to work with it.

Why test properties and test states are so important? It is these properties and these states that give us the understanding of the world. To understand the world means to have an idea for things we do not see directly. For example, let's know that the door on the third floor is unlocked. In order to formulate this proposition, we need the respective test. In order to decide whether this proposition is true, we will need a theory of the test state of this test.

Many researchers in the field of AI agree that AI should be able to make logical conclusions based on a system of automated theorem proving – for example, propositional or predicate calculus. We also share this view, but the question is how from the sequence (action, observation) to reach propositional or predicate calculus. To have a propositional calculus we need propositions. To make a predicate calculus, we will need predicates. If we cannot understand the sequence (action, observation), it would be just a noise for us. How can we make propositions and predicates from this sequence? The binding unit we need is test properties and test states. For example, the test property "The door is unlocked" may be the proposition we are looking for. The test state "Door X is unlocked" will be the predicate from which we can make statements of the type "All doors are unlocked."